\documentclass[conference,letterpaper,10pt]{IEEEtran}
\IEEEoverridecommandlockouts
\usepackage[T1]{fontenc}
\usepackage{cite}
\usepackage{amsmath,amssymb,amsfonts}
\usepackage{graphicx}
\usepackage{booktabs}
\usepackage{array}
\usepackage{xcolor}
\usepackage{colortbl}
\usepackage[hidelinks]{hyperref}
\usepackage{url}
\hypersetup{pdftitle={Conditional Rank Allocation for Taxonomy-Aware Medical Language Model Adaptation},pdfauthor={Guangyuan Dong, Ziwei Hong, Xuehao Zhou, Zidong Yu, Bingchen Liu, Kehan Liu, Chuang Liu, Rong Fu, Yuchao Hou}}
\usepackage{xspace}
\usepackage{bm}
\usepackage{balance}
\definecolor{arborteal}{HTML}{1F6E6B}
\definecolor{arborterra}{HTML}{C4785C}
\definecolor{rowours}{HTML}{DCE7D9}
\definecolor{rowhead}{HTML}{E7E2F0}
\definecolor{rowfixed}{HTML}{F0F5F9}
\definecolor{rowstructured}{HTML}{FFF8EC}
\definecolor{rowcontrol}{HTML}{F2EDF8}
\newcommand{\method}{\textsc{Arbor}\xspace}

\newcommand{\rank}{\mathrm{rank}}
\newcommand{\Diag}{\operatorname{Diag}}

\begin{document}
\title{Conditional Rank Allocation for Taxonomy-Aware Medical Language Model Adaptation}
\author{
\IEEEauthorblockN{Guangyuan Dong\textsuperscript{1,\dag,\ddag},
Ziwei Hong\textsuperscript{4,\dag},
Xuehao Zhou\textsuperscript{2,3,\dag},
Zidong Yu\textsuperscript{6},
Bingchen Liu\textsuperscript{5,\dag}\\
Kehan Liu\textsuperscript{1},
Chuang Liu,
Rong Fu,
Yuchao Hou\textsuperscript{*}}
\IEEEauthorblockA{\fontsize{9}{10.5}\selectfont
\textsuperscript{1}National University of Singapore, Singapore\\
\textsuperscript{2}Department of Urology, Shanghai General Hospital, Shanghai, China\\
\textsuperscript{3}Shanghai Jiao Tong University School of Medicine, Shanghai, China\\
\textsuperscript{4}University of Pennsylvania, USA \quad
\textsuperscript{5}Shandong University, China \quad
\textsuperscript{6}Syracuse University, USA\\
Email: guangyuan@u.nus.edu \qquad
\textsuperscript{\dag}These authors contributed equally. \quad \textsuperscript{\ddag}Project Lead. \quad \textsuperscript{*}Corresponding author.}}
\maketitle
\begin{abstract}
Medical question answering spans specialties and clinical operations that may
benefit from different adaptation directions. We propose \method, a
parameter-efficient method that selects rank-one components from a shared
low-rank basis for each question. An additive gate combines question
representations, specialty tags, operation tags, and their interaction; a
learned coefficient scales the adapter residual. An illustrative separation
under orthogonal, equiprobable subtasks shows how conditional selection can
avoid an approximation floor faced by a fixed update with the same active
rank. This result motivates the design without asserting a corresponding
bound for medical corpora. On Qwen3-8B across CMB, CMExam, MedQA, and MedMCQA,
five-seed experiments yield 69.69\% mean accuracy across benchmarks,
exceeding LoRA r16 and MoELoRA by 1.26 and 1.30 percentage points,
respectively. The reported advantage over LoRA r16 increases from 0.08 to
1.94 points as training expands from one to seven specialties. Tag
perturbations and atom masking support the usefulness of clinical routing,
while atom clusters align with the supplied specialty labels
(adjusted Rand index 0.62). Calibration, transfer, and measured costs further
characterize the method. These findings support structured conditional
adaptation for medical QA, while leaving clinical safety and broader
deployment untested.
\end{abstract}

\section{Introduction}
\label{sec:intro}

Medical question answering spans specialties and clinical operations that may
benefit from different adaptation directions. Standard LoRA applies one
low-rank update to every input~\cite{lora2022}; a drug-interaction question
and an anatomy fact therefore share the same adaptation subspace.
Conditional selection offers a way to specialize updates while reusing a
stored basis.

Specialty and clinical operation describe complementary aspects of a
question. A specialty identifies its knowledge context, while an operation
describes the requested task, such as recall, diagnosis, or treatment.
Questions within one specialty can require different operations, and the
same operation can recur across specialties. This motivates separate
routing contributions together with a pair-specific interaction, rather than
a single undifferentiated clinical label.

\method (Adaptive Rank allocation with Budgeted On-demand Routing) selects
rank-one atoms per question using an additive gate over question
representations, specialty, operation, and their interaction. A learned
coefficient scales the residual, and a discrete mask records active
components. Our contribution is this clinical structure and its evaluation,
building on existing rank-level routing rather than introducing that
mechanism.

An orthogonal-subtask construction illustrates a representational advantage
at matched active rank. Its assumptions do not imply monotonic gains for
real medical corpora or explain comparisons against rank-16 LoRA when only
seven specialties are tested. We examine these gains empirically through
heterogeneity, tag perturbations, and atom masking.

Across four Chinese and English benchmarks and five seeds, \method reaches
69.69\% mean accuracy, exceeding LoRA r16 and MoELoRA by 1.26 and 1.30
percentage points. Clustering evaluates alignment with supplied clinical
labels; calibration, transfer, and measured costs characterize the method
beyond accuracy. Active rank, parameter count, and runtime are distinguished
throughout.

\begin{figure*}[t]
\centering
\includegraphics[width=\textwidth]{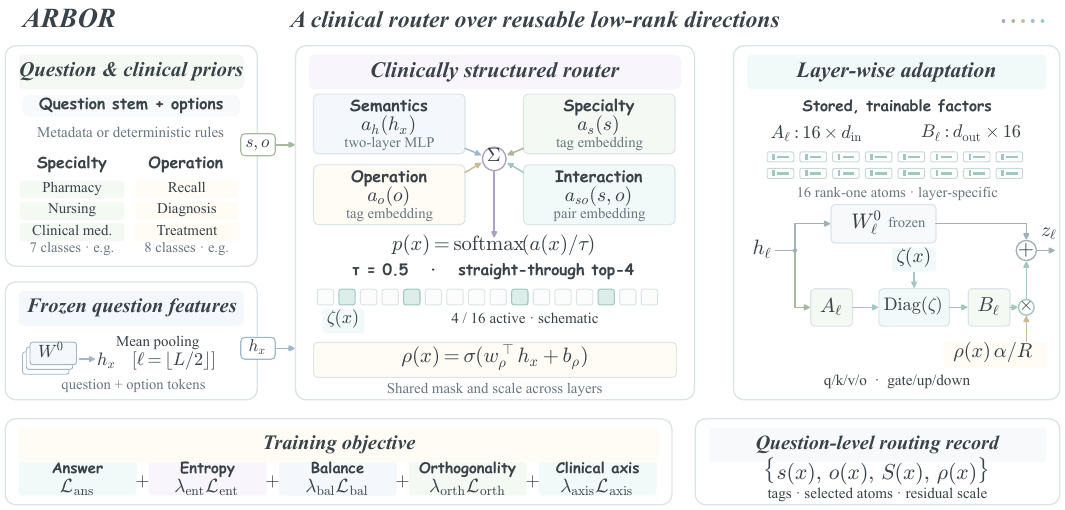}
\caption{\method overview. Clinical priors and frozen question features
feed four additive routing branches. Straight-through top-$\kappa$ selection
activates four of sixteen stored rank-one atoms; a learned coefficient scales
their contribution alongside the frozen backbone. Mask and scale are shared
across target layers. The training objective and question-level routing record
are shown below. Highlighted mask positions are schematic.}
\label{fig:arch}
\end{figure*}
\section{Related Work and Positioning}
\label{sec:related}

LoRA uses a shared low-rank update~\cite{lora2022}, while AdaLoRA reallocates
rank across modules~\cite{adalora2023}. MoELoRA routes among
adapters~\cite{moelora2024}; HydraLoRA shares a down-projection while
diversifying expert up-projections~\cite{hydralora2024}.

Biomedical post-training also includes BFT, which reweights token- and
sample-level losses for knowledge alignment~\cite{bft2025}.
Knowledge-guided pathway prototypes in scProtoTransformer provide
interpretable biological representations~\cite{scproto2026}.

Input-conditioned reuse of learned primitives also appears in
PriMD~\cite{primd2026}, which retrieves semantic primitives to compensate
for missing modalities. \method applies conditional selection to rank-one
parameter updates for medical question answering.

Arrow~\cite{arrow2024}, MeteoRA~\cite{meteora2024}, and
LoRA-Mixer~\cite{loramixer2025} compose pretrained adapter pools;
MedAdapter~\cite{medadapter2024} verifies candidate answers at test time.
These pipelines differ from adapters trained under our common CMB-source
budget. Matching active rank, stored rank, or parameter count guarantees
neither equal runtime nor identical capacity.

In biomedical named entity recognition, MITE combines code-formatted
instruction tuning with entity-level voting~\cite{mite2026}, while GAMA
combines verified annotation guidelines with schema-constrained
generation~\cite{gama2026}. \method uses clinical tags to condition parameter
updates for question answering.

SMoRA treats individual ranks as experts with straight-through
selection~\cite{smora2025}. \method adds specialty, operation, and interaction
logits plus residual scaling, and evaluates their contribution to clinical
routing and supplied-label alignment.

\section{Why Conditional Rank Allocation Helps}
\label{sec:theory}

We give an illustrative linear separation, not a medical-model risk bound.
Let $\Delta_j=B_{:,j}A_{j,:}$, $j\in[R]$. Fixed selection applies
$\Delta_S=\sum_{j\in S}\Delta_j$ to every $h(x)$; conditional selection
applies $\Delta_{S(x)}h(x)$ with $|S(x)|=\kappa$. Constant routing includes
fixed supports. Strict inclusion requires nondegenerate atoms and inputs
enabling outputs unavailable to constant routing; it need not hold for every
basis.

\paragraph{Orthogonal-subtask separation}
\label{sec:separation}
Consider $m$ equiprobable regions with representation $h(x)=v_i$ and target
$u_i$ in region $i$, where both $\{u_i\}_{i=1}^m$ and
$\{v_i\}_{i=1}^m$ are orthonormal. For a fixed linear update $\Delta$, define
\begin{equation}
\mathcal R(\Delta)=\frac{1}{m}\sum_{i=1}^{m}
  \|\Delta v_i-u_i\|_2^2.
\label{eq:toy-risk}
\end{equation}
The target map $T=\sum_i u_iv_i^{\top}$ has $m$ unit singular values.
Consequently, the Eckart--Young theorem~\cite{eckartyoung1936} gives
\begin{equation}
\inf_{\rank(\Delta)\le\kappa}\mathcal R(\Delta)
  =\max(0,1-\kappa/m).
\label{eq:toy-floor}
\end{equation}
With $R\ge m$, store $\Delta_i=u_iv_i^{\top}$ and route region $i$ to
atom $i$, yielding zero risk at active rank one. Larger budgets remain exact
if sufficient additional orthogonal atoms contribute zero on $v_i$.
This construction trades stored capacity for active rank.

\paragraph{Scope of the argument}
\label{sec:prediction}
Orthogonality, equiprobable regions, and correct routing are assumed.
Overlap, imbalance, routing errors, and nonlinear representations alter the
comparison. More medical specialties need not mean more independent target
directions. Rank-16 LoRA has zero floor in Eq.~\eqref{eq:toy-floor} for
$m\le7$, so its heterogeneity comparison is empirical, not bound validation.
Matched rank-4 sweeps and multiple specialty combinations would strengthen
the connection. A generalization analysis must also account for
input-dependent router complexity, not only support counts.

\section{Method}
\label{sec:method}

\subsection{Conditional update}
For frozen projection $W^0_\ell$ and hidden state $h_\ell$, \method computes
\begin{equation}
z_\ell=W^0_\ell h_\ell+\rho(x)\frac{\alpha}{R}
B_\ell\Diag(\zeta(x))A_\ell h_\ell,
\label{eq:arbor}
\end{equation}
where $A_\ell\in\mathbb R^{R\times d_{\mathrm{in}}}$ and
$B_\ell\in\mathbb R^{d_{\mathrm{out}}\times R}$ are trainable,
$\zeta(x)\in\{0,1\}^{R}$ has $\kappa$ active entries, and
$\rho(x)\in[0,1]$. Atom $j$ is $B_{\ell,:,j}A_{\ell,j,:}$.
The same mask and scale are used across target layers for a given question,
while each projection retains its own $A_\ell,B_\ell$.

\subsection{Clinically structured gate}
Question representation $h_x$ is the mean-pooled frozen-backbone hidden
state over question and option tokens at layer $\lfloor L/2\rfloor$.
The gate combines semantic and clinical logits:
\begin{equation}
a(x)=a_h(h_x)+a_s(s(x))+a_o(o(x))+a_{so}(s(x),o(x)).
\label{eq:gate}
\end{equation}
Here $a_h$ is a two-layer MLP; $a_s,a_o,a_{so}$ are embedding tables
for specialty, operation, and their pair. This additive form allows each
clinical axis to contribute separately while retaining pair-specific effects.
We compute
\begin{align}
p(x)&=\mathrm{softmax}(a(x)/\tau),\quad
\zeta(x)=\mathrm{ST\text{-}top\text{-}\kappa}(p(x)),
\label{eq:mask}\\
\rho(x)&=\sigma(w_\rho^{\top}h_x+b_\rho).
\label{eq:rho}
\end{align}
The forward pass selects the $\kappa$ largest entries with a binary mask;
a straight-through estimator passes gradients through soft
probabilities~\cite{ste2013,gumbel2017}. The coefficient $\rho$ learns how
strongly to inject the residual. It is neither calibrated uncertainty nor a
validated measure of clinical safety. Gate features must be obtained from the
frozen backbone before the conditioned adapter pass; their extraction or
caching is part of the implementation cost.

\subsection{Clinical tags}
Seven specialty labels and eight operation labels come from available
metadata or deterministic rules on the question and options. For example, specialty rules include drug/dose/contraindication for Pharmacy,
nursing/care plan for Nursing, and incidence/screening/vaccine for Public
Health; the fallback is Clinical Medicine. Operation rules cover diagnosis,
treatment, medication, examination, mechanism, nursing action, and prevention,
with Recall as fallback. Gold answers are excluded, and target-benchmark
labels are not used to tune rules or routing. 

Two annotators labeled a 500-item sample (Table~\ref{tab:tagval}); specialty and operation tags
achieve 91.4\% and 84.6\% accuracy, respectively.
Inter-annotator agreement is $\kappa=0.86$ and $0.79$; rule coverage is
100\% through the fallback labels. These measurements describe the sampled
items and do not establish equal tag quality across languages or datasets.

\begin{table}[t]
\centering
\caption{Clinical-prior validation on 500 annotated questions.
Agreement is inter-annotator $\kappa$.}
\label{tab:tagval}
\footnotesize
\setlength{\tabcolsep}{4.5pt}
\begin{tabular}{lcccc}
\toprule
\rowcolor{rowhead}
Axis & Accuracy & Coverage & Majority & Agreement \\
\midrule
\rowcolor{rowfixed}
Specialty (7) & 91.4\% & 100\% & 28.6\% & 0.86 \\
\rowcolor{rowours}
Operation (8) & 84.6\% & 100\% & 24.2\% & 0.79 \\
\bottomrule
\end{tabular}
\end{table}

\subsection{Training and computation}
Answer-only supervised fine-tuning uses
\begin{equation}
\begin{split}
\mathcal L=\mathcal L_{\mathrm{ans}}
&+\lambda_{\mathrm{ent}}\mathcal L_{\mathrm{ent}}
+\lambda_{\mathrm{bal}}\mathcal L_{\mathrm{bal}}\\
&+\lambda_{\mathrm{orth}}\mathcal L_{\mathrm{orth}}
+\lambda_{\mathrm{axis}}\mathcal L_{\mathrm{axis}}.
\end{split}
\label{eq:loss}
\end{equation}
Entropy and balancing terms regularize routing;
$\mathcal L_{\mathrm{orth}}=\|AA^{\top}\odot(1-I)\|_F^2$
penalizes correlation between atom directions. Decorrelation also regularizes
molecular interaction features in DSIL-DDI~\cite{dsilddi2024}.
A supervised contrastive
$\mathcal L_{\mathrm{axis}}$ encourages questions from the same specialty
to have similar routing distributions. Since specialty labels enter both
the gate and this objective, clustering evaluates alignment with supplied
labels rather than independent discovery of a taxonomy.
We use $R=16$, $\kappa=4$, $\alpha=32$, $\tau=0.5$, and regularizer weights
$(0.01,0.001,0.1,0.5)$ in the order above; remaining training settings are
in Section~\ref{sec:setup}. Rank selection costs $O(R\log\kappa)$ once per
question. Activating four atoms does not imply the parameter count or runtime
of rank-4 LoRA: the full basis is stored, and routing, feature extraction, and
the adapter implementation contribute to measured cost.

\section{Experimental Setup}
\label{sec:setup}

\paragraph{Data and evaluation}
Using Qwen3-8B~\cite{qwen3_2025}, source training uses a route-balanced
CMB~\cite{cmb2024} subset of 4,200 questions (600 per specialty), disjoint
from evaluation. Main evaluation uses Chinese CMB (4,149 questions) and
CMExam~\cite{cmexam2023}, and English MedQA~\cite{medqa2020} (700 test
questions) and MedMCQA~\cite{medmcqa2022} (700 validation questions).
PubMedQA~\cite{pubmedqa2019} provides out-of-family evaluation.
Accuracy is primary; macro is the unweighted four-benchmark average,
not pooled question accuracy. Exact evaluation manifests and the CMExam
split and sample count remain reproducibility limitations.

\paragraph{Training and statistical analysis}
Trainable methods use 400 updates, effective batch size 8, learning rate
$10^{-4}$, maximum sequence length 768, answer-only supervised fine-tuning,
greedy decoding, and a common answer parser and prompt template within each
language. \method uses $R=16$, $\kappa=4$, $\alpha=32$, $\tau=0.5$, and
$(\lambda_{\rm ent},\lambda_{\rm bal},\lambda_{\rm orth},
\lambda_{\rm axis})=(0.01,0.001,0.1,0.5)$, targeting the
q/k/v/o/gate/up/down projections. Hyperparameters were selected using a
400-question held-out source slice, without target-benchmark labels.
Runs use an A100-80GB, PyTorch 2.3, and bf16. Results are means and standard deviations over seeds $\{13,29,47,71,97\}$.
The paired differences and their 95\% confidence intervals in
Fig.~\ref{fig:main} summarize variation across training seeds; they do not
quantify uncertainty from resampling the benchmark population.

\paragraph{Comparators}
LoRA r4 matches the active rank, LoRA r16 the atom-bank size, and LoRA r24
approximately the trainable parameter count. We also compare DoRA~\cite{dora2024} r24,
HydraLoRA (four heads), and MoELoRA (four rank-8 experts, top-2).
The rank-wise control uses an r16/top-4 adapter inspired by
SMoRA~\cite{smora2025}, but removes clinical branches, their interaction,
axis contrast, orthogonality, and injection scaling. It therefore compares
the complete design with a simplified router and does not isolate the
effect of clinical tags alone. Adapter-library results are reported
separately because they compose pretrained adapters and are not matched
source-training comparisons.

\begin{table*}[t]
\centering
\caption{Qwen3-8B medical QA accuracy (\%). Trainable methods share the
CMB-source protocol; values are mean $\pm$ standard deviation over five
seeds. Macro is the unweighted benchmark average. Bold marks the best
trainable method. Adapter libraries compose pretrained adapters and are
reported separately without across-seed uncertainty.}
\label{tab:main}
\footnotesize
\setlength{\tabcolsep}{5pt}
\begin{tabular}{lcccccc}
\toprule
\rowcolor{rowhead}
Method & Trainable (M) & CMB & CMExam & MedQA & MedMCQA & Macro \\
\midrule
Frozen Qwen3-8B & -- & 71.68 & 68.95 & 62.86 & 54.86 & 64.59 \\
\rowcolor{rowfixed}
LoRA r4 & 10.91 & $77.62\pm.21$ & $75.67\pm.24$ & $62.91\pm.42$ & $57.19\pm.38$ & $68.35\pm.18$ \\
\rowcolor{rowfixed}
LoRA r16 & 43.65 & $77.79\pm.19$ & $75.98\pm.22$ & $62.60\pm.45$ & $57.34\pm.41$ & $68.43\pm.21$ \\
\rowcolor{rowfixed}
LoRA r24 & 65.47 & $77.95\pm.23$ & $75.88\pm.25$ & $62.71\pm.44$ & $57.62\pm.37$ & $68.54\pm.22$ \\
\rowcolor{rowstructured}
DoRA r24 & 66.87 & $78.48\pm.20$ & $75.86\pm.28$ & $62.19\pm.47$ & $56.90\pm.44$ & $68.36\pm.17$ \\
\rowcolor{rowstructured}
HydraLoRA & 44.90 & $78.02\pm.22$ & $75.92\pm.24$ & $63.15\pm.46$ & $57.34\pm.40$ & $68.61\pm.20$ \\
\rowcolor{rowstructured}
MoELoRA & 92.60 & $77.88\pm.24$ & $75.41\pm.26$ & $63.02\pm.51$ & $57.25\pm.42$ & $68.39\pm.20$ \\
\rowcolor{rowcontrol}
Rank-wise control & 64.88 & $78.21\pm.22$ & $75.74\pm.27$ & $63.28\pm.48$ & $56.75\pm.45$ & $68.49\pm.11$ \\
\rowcolor{rowours}
\textbf{\method} & 66.59 & $\mathbf{78.55}\pm.18$ & $\mathbf{76.71}\pm.21$ & $\mathbf{64.87}\pm.35$ & $\mathbf{58.63}\pm.33$ & $\mathbf{69.69}\pm.12$ \\
\midrule
MedAdapter~\cite{medadapter2024} & -- & 77.61 & 77.07 & 64.00 & 56.86 & 68.89 \\
Arrow~\cite{arrow2024} & -- & 76.69 & 77.30 & 61.57 & 56.57 & 68.03 \\
MeteoRA~\cite{meteora2024} & -- & 76.67 & 75.33 & 61.57 & 56.57 & 67.54 \\
LoRA-Mixer~\cite{loramixer2025} & -- & 77.03 & 75.52 & 62.00 & 57.14 & 67.92 \\
\bottomrule
\end{tabular}
\end{table*}

\section{Results and Analysis}
\label{sec:results}

\begin{figure*}[t]
\centering
\includegraphics[width=\textwidth]{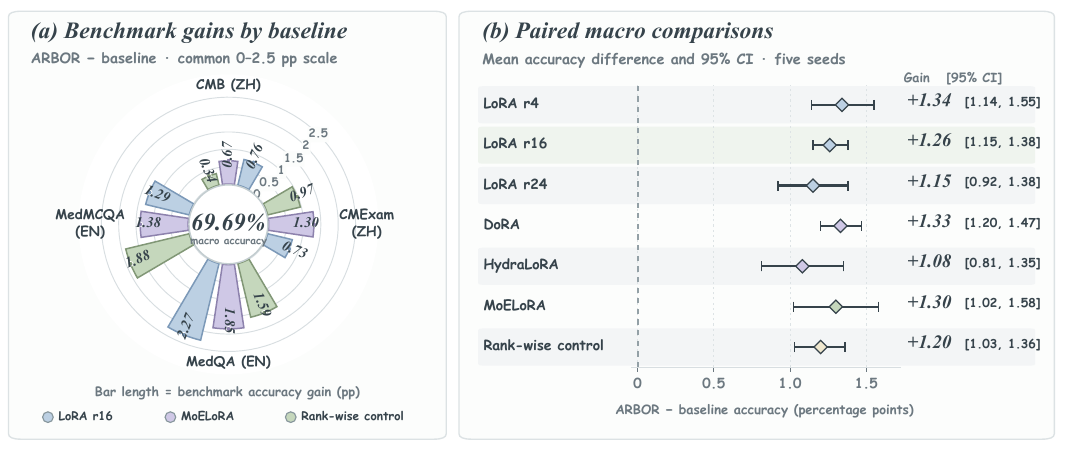}
\caption{Accuracy improvements under matched source training.
(a) Benchmark gains over three baselines share a 0--2.5 percentage-point
radial scale, with zero at the inner ring; the center reports \method's
macro accuracy.
(b) Paired macro differences against seven trainable baselines, with reported
95\% intervals across five training seeds. All intervals exclude zero.}
\label{fig:main}
\end{figure*}

\paragraph{Main results}
\method achieves $69.69\pm0.12$ macro accuracy and the highest accuracy on
each benchmark among the trainable methods in Table~\ref{tab:main}.
Its macro advantages are 1.26 percentage points over LoRA r16, 1.30 over
MoELoRA, and 1.20 over the rank-wise control. The paired across-seed
intervals are $[1.15,1.38]$, $[1.02,1.58]$, and $[1.03,1.36]$,
respectively (Fig.~\ref{fig:main}). Five seeds provide limited information about training variability;
these intervals do not include uncertainty from evaluation-set sampling. Adapter libraries sometimes outperform \method on individual
benchmarks: Arrow reaches 77.30 on CMExam, compared with 76.71 for \method.

\paragraph{Transfer}
Relative to LoRA r16, the gains are 0.76 and 0.73 points on the Chinese
CMB and CMExam sets, and 2.27 and 1.29 points on the English MedQA and
MedMCQA sets. Thus, the improvement extends beyond the training language.
On held-out PubMedQA, \method reaches $74.15\pm0.30$, compared with the
reported strongest baseline at 72.61. These benchmark results suggest
transfer, without establishing performance on open-ended clinical tasks.

\begin{figure*}[t]
\centering
\includegraphics[width=\textwidth]{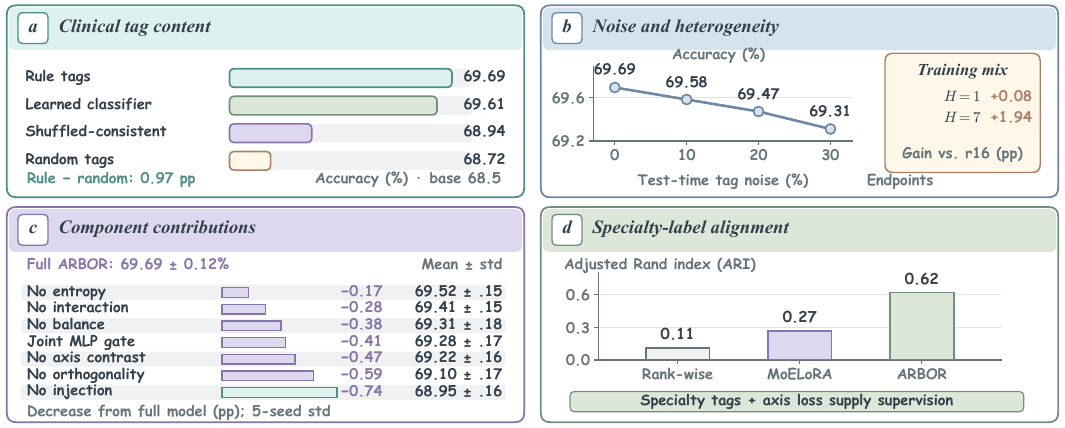}
\caption{Clinical routing and component analysis.
(a) Tag-source controls with other components held fixed.
(b) Aggregate accuracy under test-time tag noise (25 runs per level), with
the reported training-heterogeneity endpoints shown separately.
(c) Component ablations relative to full \method; annotation uncertainty
denotes standard deviation across five seeds.
(d) Alignment with supplied specialty labels.
Only available numerical summaries are shown; heterogeneity endpoints are
not interpolated.}
\label{fig:causal}
\label{fig:heterogeneity}
\end{figure*}

\subsection{Routing and Clinical Structure}
\label{sec:mechanism}

\paragraph{Training heterogeneity}
We vary the number of training specialties over
$H\in\{1,2,3,4,5,7\}$, with 4,200 training questions, a fixed mixed
evaluation set, and five seeds per setting. The reported \method--LoRA r16 advantage increases from 0.08 points at $H=1$ to 1.94 at $H=7$.
Figure~\ref{fig:heterogeneity}(b) shows these endpoints. The trend is an
empirical observation rather than a direct test of the rank-$\kappa$
separation in Section~\ref{sec:theory}: LoRA r16 has rank greater than
the maximum seven specialties. The experiment does not establish robustness to different specialty
combinations. A rank-matched LoRA r4 sweep and repeated combinations would
strengthen the interpretation; low-$H$ sampling and evaluation aggregation
also require fuller documentation.

\paragraph{Tag interventions}
Rule-derived tags give macro accuracy 69.69; a learned classifier gives
69.61, shuffled-but-consistent tags 68.94, and random tags 68.72
(Fig.~\ref{fig:causal}(a)). Randomization therefore removes 0.97 of the
1.20-point advantage over the rank-wise control. These interventions
support the relevance of tag content within the tested model; they do not
establish a general causal claim about clinical reasoning.
Under 0\%, 10\%, 20\%, and 30\% test-time tag noise, aggregate accuracy is
69.69, 69.58, 69.47, and 69.31, respectively. The aggregate trend supports sensitivity to tag quality. Subgroup effects
require a defined clean-subset reference and are omitted from the figure.

\paragraph{Atom specialization and taxonomy alignment}
Atom analysis indicates shared and specialty-selective activation and
masking effects. Clustering by active-atom signatures gives
adjusted Rand index 0.62 against specialty labels (Fig.~\ref{fig:causal}(d)), compared with 0.27 for
MoELoRA and 0.11 for the rank-wise control. This is alignment with the
supplied taxonomy: specialty tags enter the router, and the axis loss
explicitly promotes within-specialty agreement. The result cannot establish
independent discovery of a medical taxonomy. Taxonomy metrics without specialty inputs and axis supervision remain
an open control. Activation patching and pathway-level interventions have been
used to investigate hallucination mechanisms in vision-language
models~\cite{dualpathway2026}. Our atom-masking analysis measures sensitivity
to adapter components, without establishing causal explanations of clinical
knowledge.

\paragraph{Component ablations}
Figure~\ref{fig:causal}(c) reports the component ablations. Removing the
interaction costs 0.28 points, while removing axis contrast costs 0.47;
the additive gate exceeds a joint MLP by 0.41. Removing injection scaling
has the largest reported loss, 0.74 points. These results support the full
design, but specialty-only and operation-only comparisons are unavailable.
The reported budget sweep peaks at $\kappa=4$, without implying a
universal optimal rank budget.

\paragraph{Injection strength}
Mean $\rho(x)$ is 0.55 for recall and 0.83 for diagnosis. It is 0.84 on
questions where \method corrects a base-model error and 0.61 where the base
is already correct. These associations describe input-dependent update strength;
$\rho$ is not a validated uncertainty or safety measure. Reported ECE is
4.2 for \method and 6.4 for LoRA r16~\cite{calibration2017}; interpreting
this difference requires a fully documented confidence and binning protocol.

\section{Transfer and Computational Cost}
\label{sec:generalization}

\paragraph{Source and backbone checks}
When the Chinese training source changes to CMExam, \method obtains
70.11 macro accuracy versus 69.02 for MoELoRA. Reported five-seed
advantages over MoELoRA are 0.97 points on Qwen3-4B and 0.84 on
Llama-3.1-8B~\cite{llama3_2024}. These checks extend the evaluation
across two model families and scales up to 8B; they do not establish
scaling behavior beyond that range.

\begin{table}[t]
\centering
\caption{Reported Qwen3-8B costs on one A100-80GB. Parameters are total
trainable parameters; memory is peak inference memory. Latency is per
question under greedy decoding.}
\label{tab:efficiency}
\footnotesize
\setlength{\tabcolsep}{4pt}
\begin{tabular}{lccccc}
\toprule
\rowcolor{rowhead}
Method & Params & Active & Train & Latency & Memory \\
\rowcolor{rowhead}
 & (M) & rank & (min) & (ms) & (GB) \\
\midrule
\rowcolor{rowfixed}
LoRA r4 & 10.91 & 4 & 38 & 416 & 19.6 \\
\rowcolor{rowfixed}
LoRA r16 & 43.65 & 16 & 41 & 418 & 19.8 \\
\rowcolor{rowfixed}
LoRA r24 & 65.47 & 24 & 44 & 421 & 20.0 \\
\rowcolor{rowstructured}
DoRA & 66.87 & 24 & 58 & 435 & 21.0 \\
\rowcolor{rowstructured}
HydraLoRA & 44.90 & 16 & 49 & 452 & 20.6 \\
\rowcolor{rowstructured}
MoELoRA & 92.60 & 16 & 63 & 486 & 22.4 \\
\rowcolor{rowcontrol}
Rank-wise control & 64.88 & 4 & 46 & 428 & 20.1 \\
\rowcolor{rowours}
\method & 66.59 & 4 & 47 & 431 & 20.1 \\
\bottomrule
\end{tabular}
\end{table}

\paragraph{Active rank and measured cost}
\method activates four atoms per question but stores a 16-atom bank and
a router. Table~\ref{tab:efficiency} reports 66.59M trainable parameters,
47 minutes of training, 431 ms per question, and 20.1 GB peak inference
memory. Compared with MoELoRA, the total parameter count is 28.1\% lower,
and the reported training and inference costs are smaller. Relative to LoRA
r4, however, \method uses more parameters and memory and has 3.6\% higher
latency. Dense masking does not translate a four-atom budget into the
execution cost of a rank-4 adapter.

The router uses features from a frozen middle-layer pass before adaptation.
The timing protocol does not specify feature caching or whether this pass
is included, and parameter counts do not separate the atom bank and router.
These implementation details, along with generation lengths and batching,
limit conclusions about end-to-end serving cost.

\section{Discussion and Limitations}
\label{sec:discussion}

The results support clinically structured conditional rank allocation in
the evaluated medical QA settings. The illustrative separation assumes
orthogonal, equiprobable targets, whereas clinical specialties overlap and
may be imbalanced. Taxonomy alignment is conditioned on supplied specialty
priors and axis supervision.
\balance

Open controls include a LoRA r4 heterogeneity sweep, repeated specialty
combinations, single-axis gates, and clinical-input ablations retaining
orthogonality and scaling. Full reproducibility requires evaluation
identifiers, bilingual tagging rules, router dimensions, annotation
sampling, and timing scope. Seed intervals do not replace prediction-level
uncertainty. Public QA benchmarks and backbones up to 8B leave open-ended
reasoning, prospective clinical validity, and larger-model behavior
untested. Selected atoms and injection strength are not validated clinical
explanations or safety signals.

\section{Conclusion}
\label{sec:conclusion}

\method combines question semantics, specialty and operation priors,
top-$\kappa$ rank selection, and learned injection scaling in a shared atom
bank. Across five seeds, it improves four-benchmark macro accuracy over
fixed-rank and expert-based baselines. Tag interventions and ablations
support the structured design, while the heterogeneity trend motivates
rank-matched experiments. The orthogonal-task analysis explains a potential
benefit of conditional allocation within its assumptions.

\bibliographystyle{IEEEtran}
\bibliography{references}
\end{document}